\documentclass{article}

     \PassOptionsToPackage{numbers, compress}{natbib}

\usepackage[final]{neurips_2024}

\usepackage[utf8]{inputenc} 
\usepackage[T1]{fontenc}    
\usepackage{hyperref}       
\usepackage{url}            
\usepackage{booktabs}       
\usepackage{amsfonts}       
\usepackage{nicefrac}       
\usepackage{microtype}      
\usepackage{xcolor}         
\usepackage{subfig}
\usepackage{float}
\usepackage{graphicx} 
\usepackage{amsmath}
\usepackage{makecell}
\usepackage{multirow}
\usepackage{hyperref}

\title{Prune and Repaint: Content-Aware Image Retargeting for any Ratio}

\begin{document}


%
\def\thefootnote{$*$}\footnotetext{Corresponding author (\text{haochen303@seu.edu.cn)}}
\author{
  \begin{tabular}[t]{c}
   Feihong Shen$^{1,2,4}$ \\
    \texttt{feihongshen@seu.edu.cn}
  \end{tabular} \hspace{1em}
  \begin{tabular}[t]{c}
    Chao Li$^{2}$ \\
    \texttt{lllcho.lc@alibaba-inc.com}
  \end{tabular} \vspace{0.6em}\\ 
  \begin{tabular}[t]{c}
    \textbf{Yifeng Geng}$^{2}$ \\
     \texttt{cangyu.gyf@alibaba-inc.com}
  \end{tabular} \hspace{0.2em}
  \begin{tabular}[t]{c}
    \textbf{Yongjian Deng}$^{3}$ \\
     \texttt{yjdeng@bjut.edu.cn}
  \end{tabular} \hspace{0.2em}
  \begin{tabular}[t]{c}
    \textbf{Hao Chen}$^{*1,4}$ \\
    \texttt{haochen303@seu.edu.cn}
  \end{tabular} \\
  \\
 \begin{tabular}[t]{c}
        $^1$\text{Southeast University} \\
  \end{tabular} \hspace{0.4em}
  \begin{tabular}[t]{c}
    $^2$\text{Alibaba Group} \\
  \end{tabular} \hspace{0.4em}
  \begin{tabular}[t]{c}
    $^3$\text{Beijing University
of Technology} \\
  \end{tabular} 
  \vspace{0.6em} \\
  \begin{tabular}[t]{c}
    $^4$\text{Key Laboratory of New Generation Artificial Intelligence Technology} \\
  \end{tabular}
}

\maketitle

\begin{abstract}

Image retargeting is the task of adjusting the aspect ratio of images to suit different display devices or presentation environments. However, existing retargeting methods often struggle to balance the preservation of key semantics and image quality, resulting in either deformation or loss of important objects, or the introduction of local artifacts such as discontinuous pixels and inconsistent regenerated content.
To address these issues, we propose a content-aware retargeting method called PruneRepaint. It incorporates semantic importance for each pixel to guide the identification of regions that need to be pruned or preserved in order to maintain key semantics. Additionally, we introduce an adaptive repainting module that selects image regions for repainting based on the distribution of pruned pixels and the proportion between foreground size and target aspect ratio, thus achieving local smoothness after pruning.
By focusing on the content and structure of the foreground, our PruneRepaint approach adaptively avoids key content loss and deformation, while effectively mitigating artifacts with local repainting. We conduct experiments on the public RetargetMe benchmark and demonstrate through objective experimental results and subjective user studies that our method outperforms previous approaches in terms of preserving semantics and aesthetics, as well as better generalization across diverse aspect ratios. 
Codes will be available at \href{https://github.com/fhshen2022/PruneRepaint}{https://github.com/fhshen2022/PruneRepaint}.

\end{abstract}

\section{Introduction}\label{section:introduction}

With the popularity of multi-screen and multi-aspect environments, people's demands for the adaptability and aesthetics of images across different devices are increasing. Consequently, image retargeting \cite{vaquero2010survey,ma2012image,fan2024comprehensive}, which aims to adjust the aspect ratio to fit various display devices or presentation environments while preserving the key content and maintaining the quality of the images, has distinctive applications yet is understudied.

\begin{figure}
	\centering
	\includegraphics[width=0.999\linewidth]{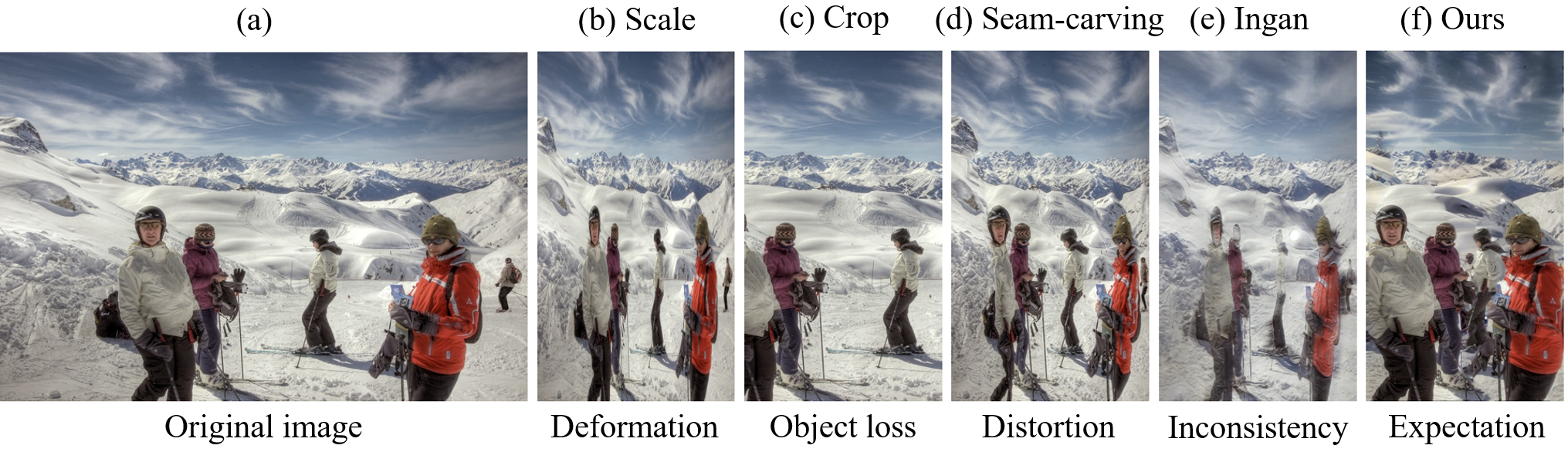}
    	\caption{ An example to show bad cases such as deformation, content loss, discontinuity in lines, inconsistent results and a good case.  }
	\label{fig:example}
\end{figure}

The core challenge of image retargeting lies in simultaneously 1) preserving the main information and 2) avoiding artifacts such as deformation and distortion on key objects. 
Intuitive solutions for this task include scaling and cropping. As shown in Figure \ref{fig:example} (b), scaling entirely preserves all contents but results in severe deformation, decreasing aesthetic appeal and image quality, making it difficult to recognize the figures. 
On the contrary, crop-based methods \cite{zhang2005auto,santella2006gaze} introduce no artifacts but often results in the loss of key semantics (see Figure \ref{fig:example} (c)). 
To relieve these problems, following methods typically use pixel-shifting operators. Noticeable works include seam-carving \cite{avidan2007seam,rubinstein2008improved}, which calculates energy using manual operators \cite{duda1973pattern,canny1986computational} to identify seams for deletion.
Some other works \cite{dong2009optimized,zhou2020weakly,kajiura2020self} further integrate these traditional operators to enhance the generalization to different scenarios.
However, as illustrated in Figure \ref{fig:example} (d), without semantics guidance for crucial regions, this line of methods often leads to content loss or distortion on important objects, as well as inconsistent pixels in the foreground.

Regarding the power of deep learning tools, some methods \cite{lin2019deepir,valdez2021fast,dickman2023smart} integrate deep semantic features to guide the deletion or preservation of pixels in traditional pixel-shifting methods. 
However, these methods fail to differentiate the semantic significance within objects (e.g., face is more important than hair in a person), thus often leading to object distortion when meets oversized objects. Moreover, these methods only focus on foreground regions, typically leading to discontinuous backgrounds and decreased aesthetic appeal. 
Towards better aesthetics and region consistency, other works \cite{cho2017weakly,shocher2019ingan,mei2021deep,fan2021unsupervised} achieve image retargeting from a generative perspective with Generative Adversarial Networks (GANs) \cite{goodfellow2020generative}. These methods often implicitly learn the semantic distribution of images to regenerate retargeted images. Due to the absence of explicit semantic prior and the weakness of GANs in capturing the global data distribution \cite{metz2016unrolled}, these methods will generate all regions without selection, resulting in inconsistent generation of key objects (Figure \ref{fig:example} (e)).

To tackle the issues mentioned above, we present a content-aware retargeting model that can maintain the essential semantics, their appearance consistency, and aesthetics while being adaptable to any aspect ratio.
To alleviate semantic loss, we introduce content-aware seam-carving (CSC), which incorporates hierarchical semantic information induced from semantic/spatial saliency to differentiate the energy to perform scene-level (i.e., background and foreground) and object-level (i.e., components in the foreground) pruning, thereby maximizing the preservation of key objects and their discriminative semantic elements. 
To mitigate artifacts introduced by pixel removal, we further propose an adaptive repainting (AR) method based on diffusion models, consisting of an Adaptive Repainting Region Determination (ARRD) module and an Image-guided Repainting (IR) module.

The two modules work together to adaptively repaint scenes with varying foreground sizes. The ARDD module is responsible for determining which regions of the image need to be repainted. It does this by identifying abrupt pixels that have a high density of removed surrounding pixels. Then, it considers the foreground size and desired ratio to determine inpainting or outpainting. This approach ensures that important objects are preserved in the image even if they exceed the expected size, resulting in a flexible method that can handle images with any aspect ratio.
Subsequently, IR refines the repainting process by using the original image as a reference to restore and repaint these regions.
Compared to previous global generation methods \cite{cho2017weakly,shocher2019ingan,mei2021deep} without maintaining foreground consistency with the original image, our approach selectively regenerates the abrupt pixels, preserving the foreground consistency and local smoothness effectively.

Our contributions can be summarized as follows:

1) We introduce a content-aware image retargeting framework that is applicable to any aspect ratio. By incorporating content-aware seam-carving, our approach enables pixel pruning with hierarchical semantic differentiation. 

2) We propose an adaptive repainting method that utilizes image-conditioned stable diffusion models. This method dynamically determines whether to inpaint or outpaint based on different aspect ratios, leading to local smoothness and aesthetically pleasing outcomes.

3) Through extensive experiments involving various aspect ratios, our method demonstrates superior performance compared to other approaches in terms of both objective and subjective evaluations. It excels in preserving object completeness, coherence, and generalization.

\section{Related Work}
\subsection{Image Retargeting}

Existing image retargeting methods revolve around two main themes: preserving the main information and avoiding artifacts. Early image retargeting methods often fail to balance these two aspects. 
For instance, scaling attempts to maintain overall elements by uniformly removing pixels but struggles with significant changes in aspect ratios, resulting in severe deformation of key objects. Cropping-based methods \cite{zhang2005auto,santella2006gaze} chooses the best window of target size from the original image, which preserves the structure but leads to the loss of crucial information outside the window. 
Seam-carving \cite{avidan2007seam} attempts to balance content completeness and quality by calculating energy maps to remove lower-energy seams. However, due to the lack of semantics, when the background is complex, this method usually result in the distortion in foreground. 

The rise of deep learning \cite{lecun1998gradient} has introduced semantic information to image retargeting. DeepIR \cite{lin2019deepir} adopts pretrained VGG \cite{simonyan2014very} to explicitly extract semantic information and retargets the image from a coarse semantic space to fine pixel space.
SmartScale \cite{dickman2023smart} utilizes existing object detection model to assist seam-carving. 
However, these methods ignore the semantic differences within important regions, resulting in deformation within the oversized regions. In addition, neglecting the background can lead to discontinuities in background pixels, thereby affecting the aesthetic appeal of the image. For aesthetics and local smoothness, some methods adopt Generative Adversarial Networks (GANs) \cite{goodfellow2020generative} to generate the retargeted results. InGAN \cite{shocher2019ingan} and SinGAN \cite{shaham2019singan} divide the image into patches and learn the internal distribution of patches, destroying the overall semantics. To training a GAN without partitioning the image, MRGAN \cite{mei2021deep} adopts multi-operator to generate a paired dataset, which is constrained by the handcraft, MCGAN \cite{dy2023mcgan} introduces mask to highlight importance areas. However, due to the limitations of implicit semantic expression, these methods preserve the global semantics but destroy the details, resulting in inconsistent appearance with the original image. 

In contrast, our method is content-aware for selective pruning and adaptive repainting. It is able to maintain the key semantics and appearance while ensuring local smoothness and aesthetics, and it has stronger generalization for different aspect ratios.

\subsection{Diffusion Models for Image Generation}
Nowadays, diffusion models \cite{ho2020denoising,song2020denoising,rombach2022high} have become the mainstream models for generative tasks due to their powerful ability to model complex distributions. Stable Diffusion \cite{rombach2022high} is the first generative model based on latent diffusion models. The progressively denoising diffusion in latent space significantly enhances the efficiency, stability, realism, and controllability of image generation. Subsequently, various improvements \cite{podell2023sdxl} and variations \cite{zhang2023adding,ye2023ip} of stable diffusion models have been proposed. For instance, SDXL \cite{podell2023sdxl} adopts a larger backbone and finetunes it using a complicated dataset with multiple aspect ratios to improve its versatility.

However, such text-to-image (T2I) models are hard to generate complex scenes and achieve more detailed control, as a significant amount of text control is labor-intensive and the T2I models struggle to accurately comprehend numerous and complex text prompts. To tackle this issue, other conditioning methods \cite{zhang2023adding,ye2023ip} are proposed. The introduction of ControlNet \cite{zhang2023adding} expands the applications of Stable Diffusion with different image-based conditional control, including depth images, mask images, canny images, etc. IP-Adapter \cite{ye2023ip}, a newly proposed image-to-image (I2I) model, introduces image prompts to control condition with an decoupled cross-attention adapter branch, highly enhancing the controllability of the generative image.

In our task, we introduce image-guided local repainting into image retargeting, which enjoys the advantage of more precise semantic preservation and more controllable local generation compared to global regeneration.

\section{Method}\label{section:method}

%
\begin{figure}
	\centering
	\includegraphics[width=0.999\linewidth]{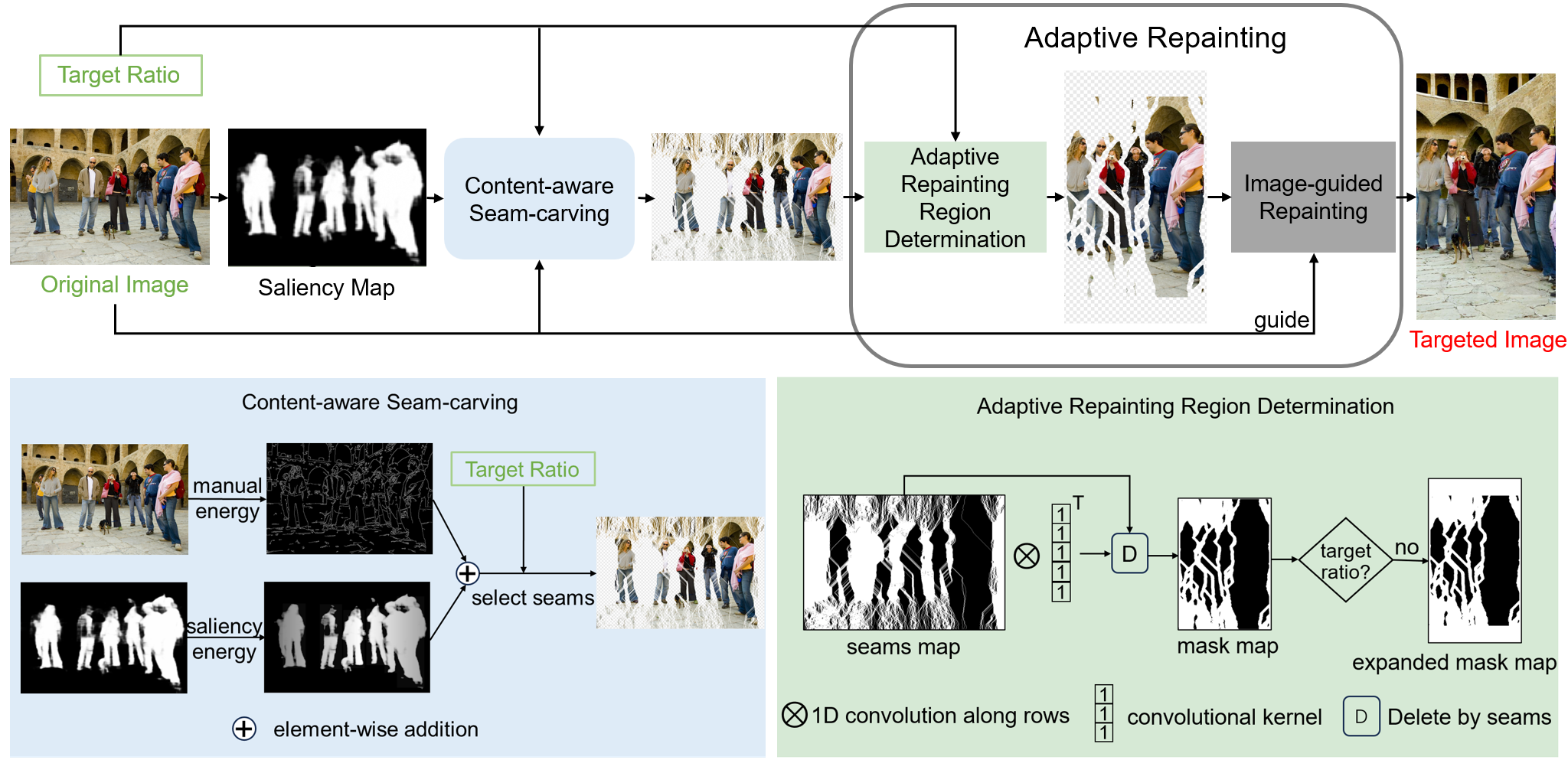}
    	\caption{The overall architecture of our proposed PruneRepaint.  The input consists of a RGB image and a target ratio. The saliency map, obtained by saliency detection, is further introduced into content-aware seam-carving module for preliminary retargeting. The preliminary retargeted result is then processed by the adaptive repainting region determination module to identify the abrupt pixel regions that need to be repainted. Utilizing  the original image as guidance, the image is inpainted with the image-guided repainting module, generating the final targeted image with target ratio. }
	\label{fig:model}
\end{figure}

\subsection{Overall Architecture}
The overall architecture is illustrated in Figure \ref{fig:model}. Specifically, a saliency detection model is adopted to obtain the semantic saliency, which will further be combined with the initial energy map to guide the determination of pruning pixels, making the pruning content-aware. After that, an adaptive repainting region determination module is applied to identify the abrupt pixels and determine the repainting regions, and an image-guided stable repainting module is further used to repaint them to output the final retargeted image.

\subsection{Content-aware Seam-carving}\label{section:csc}
Seam-carving is a typical pixel-shifting retargeting method, which calculates the energy of image and prioritizes deleting the seams with lower energy. For simplicity, we only discuss the scenario of deleting vertical seams in this section. The energy function in seam-carving is formulated as follows\cite{avidan2007seam}:
\begin{equation}\label{sc}
Energy(I(x,y)) = |\frac{\partial}{\partial x}I(x,y)|+|\frac{\partial}{\partial   y}I(x,y)|,
\end{equation}
where $I(x,y)$ denotes the pixel at position (x,y) in the image. Seam-carving is often criticized for its lack of semantic information, which can lead to the distortion of key objects. To address this issue, we propose content-aware seam-carving (CSC), which incorporates semantic and spatial saliency priors. As illustrated by the blue region in Figure \ref{fig:model}, the energy function of semantic seam-carving is formulated as follows:
\begin{equation}\label{eq2}
Energy(I(x,y)) = |\frac{\partial}{\partial x}I(x,y)|+|\frac{\partial}{\partial y}I(x,y)|+S(x,y)\odot(1-\frac{|x-x_0|}{W}),
\end{equation}
where $\odot$ denotes element-wise multiplication, $S(x,y)$ represents the saliency value at position (x,y), $x_0$ is the x-coordinate of the saliency centroid achieved by averaging the x-coordinates of all salient pixels in the saliency map, $W$ is the width of the image. The saliency map can be obtained through a pretrained salient object detection network \cite{liu2021visual}. By enhancing the energy in important areas, the saliency prior $S(x,y)$ prevents key objects from deformation. The spatial prior $(1-\frac{|x-x_0|}{W})$ further differentiates the importance within key regions, where the significance gradually decreases from the centroid towards the edges, thereby encouraging the model to prioritize seam removal from outer regions and retain key semantic elements for an object, ultimately avoiding distortion.
 
For convenience, we follow \cite{avidan2007seam} to describe a vertical seam in an image as $s^x=\{s_i^x\}_{i=1}^{W}=\{(x(i),i)\}_{i=1}^{W} $, where $x(\cdot)$ is a mapping subject to $|x(i)-x(i-1)|\leq 1$. Given the energy function, we define the cost of a seam as $cost(s) = \sum_{i=1}^{W} Energy(s_i)$. The seam to be deleted is selected by minimizing the cost:
\begin{equation}\label{eq3}
s^* = min_s \sum_{i=1}^{W}Energy(s_i).
\end{equation}
Using dynamic programming, we can efficiently find the seams with the least energy. 
We set a tolerable saliency loss ratio $\lambda$ to control the maximum loss of salient regions, which will be elaborated in Section \ref{section:arrd}. The maximun number of deleted seams is determined jointly by the saliency map and the tolerable  saliency loss ratio $\lambda$. Specifically, the quantity of seams to be deleted, which intersect the saliency map, must not exceed the product of the saliency width $W_s$ and the tolerable saliency loss ratio $\lambda$, whether the image reaches the target ratio.
We further get a binary mask $S$ where $0$ represents the low energy pixel to delete and $1$ is the pixel to be preserved. The initial retargeting results can be obtained by performing a dot product between the original image and the mask $S$ and then  concatenating the non-zero pixel regions.

\subsection{Adaptive Repainting}\label{section:repainting}
The pixel-shift method inherently introduces pixel inconsistency, and bridging the resulting pixel gap poses a significant challenge. To address this issue, we introduce Adaptive Repainting (AR), a novel approach consisting of two primary components: the Adaptive Repainting Region Determination module (ARRD) and the Image-guided Repainting module (IR).

\subsubsection{Adaptive Repainting Region Determination}\label{section:arrd}

The ARRD is designed to dynamically identify regions that require inpainting, which are characterized by inconsistencies among individual pixels. Additionally, ARRD determines the optimal repaint strategy (\textit{i.e.}, inpaint or outpaint) and corresponding regions by comparing the current ratio with the target ratio.
 
To generate the inpainting mask, we identify pixels with a high number of deleted neighboring pixels in the content-aware seam carving (CSC) result as abrupt. As depicted by the green region in Figure \ref{fig:model}, we employ a one-dimensional sliding window of length $l$ on the mask map $S$ of seam-carving to calculate the mean value within the window. This can be formalized as a one-dimensional convolution:
$M = conv1d(S,K)/l,$
where $conv1d$ is a one-dimension convolution operator, k is a one-dimensional convolution kernel of length $l$ with all values equal to 1. We then binarize $M$ into $\hat{M}$ using a threshold $\eta$, where 0 indicates areas to be inpainted and 1 denotes pixels to be preserved.

To generate the outpainting mask, we binarize the saliency map $S$ into $\hat{S}$ using the mean value as the threshold. For each connected region in the saliency map, we compute its maximum width and then take the union of all these widths to obtain the saliency width $W_s$. This can be formalized as:
\begin{equation}\label{eq6}
W_s = sum(Union(w_1,...,w_H)),
\end{equation}
where $w_i$ is the \textit{i}-th row of the binary saliency map $\hat{S}$, $Union(a,b)$ represents the union of two binary vectors $a$ and $b$, and $sum(a)$ denotes the sum of all elements in the vector $a$.
Given the target ratio $r$, we compare it with the target width $W_t$, which can be calculated as:
$W_t = H*r,$
where $H$ is the height of original image. The final targeted width $W_f$ can be determined as:
\begin{equation}\label{eq8}
W_f=\left\{
\begin{aligned}
&W_s*(1-\lambda), & W_s*(1-\lambda)> W_t \\
&W_t, & W_s*(1-\lambda)\leq W_t
\end{aligned}
\right.
,
\end{equation}
where $\lambda$ is the tolerable saliency loss ratio, set to 0.3 in our experiment. The final height $H_f$ is then calculated as $ W_f/r$, and we can determine if the image needs expanding by comparing $H_f$ and $H$. The expanded height $(H_f-h)$ will be evenly distributed to the top and bottom of the image. The outpainting mask in this stage can be merged into the inpainting mask $\hat{M}$, hence the retargeting results can be obtained with a unified repainting process with $\hat{M}$.

\begin{figure}
	\centering
    	\includegraphics[width=0.67\linewidth]{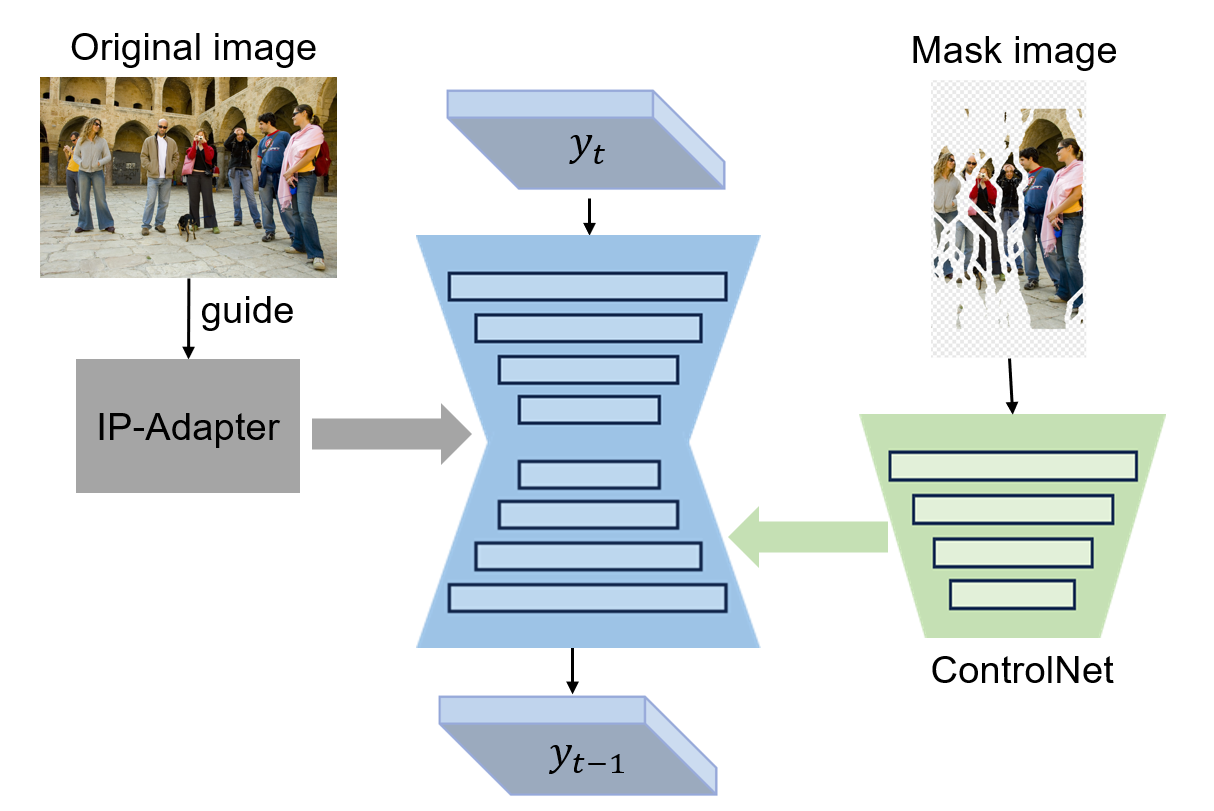}
    	\caption{The architecture of the image-guided repainting module. }
	\label{fig:igr}
\end{figure}

\subsubsection{Image-guided Repainting}\label{section:ir}

As shown in Figure \ref{fig:igr}, to achieve repainting, a pretrained ControlNet \cite{zhang2023adding}, replicated from the Stable Diffusion (SD) \cite{rombach2022high} Unet, is parallelly combined with the SD model.
This ControlNet (specifically, the inpaint version) serves to introduce features associated with visible regions of the image to be repainted. To harness the guidance provided by the original image, we further introduce an IP-Adapter \cite{ye2023ip}, which consists of a CLIP image encoder \cite{radford2021learning} and a lightweight adapter \cite{houlsby2019parameter}, to fuse image prompts with text prompts using decoupled cross-attention.

With the repainting mask $\hat{M}$ obtained in Section \ref{section:arrd}, we can formulate one reverse step in the diffusion process \cite{lugmayr2022repaint} to achieve repainting as follows:
\begin{equation}\label{eq5}
y_{t-1} = \hat{M}\odot y_{t-1}^{known}+(1-\hat{M})\odot y_{t-1}^{unknown},
\end{equation}
where $y_{t-1}^{known}$ is sampled with the unmasked pixels in the given image $\hat{M}\odot y_0$, while $y_{t-1}^{unknown}$ is sampled from the model with the previous iteration $y_t$.

\section{Experiments}

\subsection{Dataset and Evaluation Metrics}

We evaluate the proposed method on the public image retargeting datasets, RetargetMe \cite{Rubinstein2010ACS}, which contains 80 images from various scenes. According to the common sizes of prevalent electronic devices, we set the target aspect ratio for image retargeting as 16:9, 1:1, 4:3 and 9:16. 

The metrics for image retargeting have remained undetermined and existing evaluation metrics \cite{manjunath2001color,liu2011image,hsu2014objective} exhibit discrepancies with human perception, such as treating foreground and background equally. 
To intuitively evaluate the effectiveness of image retargeting methods, we propose \textbf{Saliency Discard Ratio} (SDR) to assess the semantic preservation.
The SDR can be calculated as follows:
\begin{equation}\label{eq9}
SDR = \frac{W_s^{ori}-W_s^{out}}{W_s^{ori}},
\end{equation}
where $W_s^{ori}$ is the saliency width of the original image defined in equation \ref{eq6} and $W_s^{out}$ is the saliency width of the retargeted image. 

\textbf{User study metric.} Given the subjective nature of retargeting results, we employ manual scoring as an additional evaluation method. Specifically, we invite 20 volunteers to rate the results on a scale from 0 to 3 across four aspects: content completeness, deformation, local smoothness, and aesthetics. These aspects are defined as follows: content completeness assesses whether key areas are cropped, deformation examines the degree of deformation within crucial areas, local smoothness evaluates the continuity of local regions in the image, and aesthetics evaluates the overall harmony and aesthetic appeal of the visual composition. A higher score indicates better performance.

\subsection{Implement Details}\label{detail}

Our method is implemented using Pytorch on a RTX 3090. The length of the sliding window in Section \ref{section:arrd} is set to $l=25$, and the threshold is set to $\eta=15$. We utilize the VST model \cite{liu2021visual} for salient object detection in CSC. For the image-to-image repainting model in AR, we employ a composition of SD1.5\footnote{\url{https://huggingface.co/runwayml/stable-diffusion-v1-5}}, ControlNet-Inpainting\footnote{\url{https://huggingface.co/lllyasviel/control_v11p_sd15_inpaint}} and IP-Adapter \cite{ye2023ip}.

\subsection{Compare with Other Retargeting Methods}

We quantitatively evaluate the performance of our proposed model by comparing it with three other prevalent image retargeting methods, namely scaling, cropping, seam-carving \cite{avidan2007seam}, InGAN \cite{shocher2019ingan} and full repainting which repaints the whole image with SD1.5 and IP-Adapter \cite{ye2023ip}, using the objective metric `SDR' and four subjective metrics across different aspect ratios.

\begin{table*}[ht]
	\centering
	\caption{Comparison of SDR values with other retargeting methods on the RetargetMe dataset with different aspect ratios. Lower values indicate better semantic completeness. The best results are highlighted in \textbf{bold}.} 
	\scalebox{0.93}{
	\setlength{\tabcolsep}{1.5mm}{
		\begin{tabular}{@{}l|cc|cc|cc|cc @{}}
			\hline
			Aspect Ratio & 16/9&& 4/3 && 1/1 && 9/16 \\
			\hline
			Scale           & 0.571  &&0.446 &&0.307  &&0.222  \\
			Crop           &   0.386 &&0.259 &&0.129  && 0.094  \\
			Seam-carving    &  0.490 &&0.367  &&0.242  &&0.161  \\	
                 InGAN           & 0.569  &&0.442  &&0.263   && 0.222  \\
                 FR           &0.524    &&0.423   &&0.294    &&0.214    \\
                Ours           & \textbf{0.151} &&\textbf{0.074}  &&\textbf{0.031}  &&\textbf{0.006}   \\
   
			\midrule
	\end{tabular}}
	}
	\label{tab:1}
\end{table*}

Table \ref{tab:1} presents the performance of various methods on objective metric. As shown in the table, our method achieves a significant reduction in the loss of salient regions, primarily due to the incorporation of saliency priors.

\begin{table*}[ht]
	\centering
	\caption{Subjective comparison with other retargeting methods in aspect ratio 16:9. $\uparrow$ indicates that larger are better. The best results are highlighted in \textbf{bold}.} 
	\scalebox{0.93}{
	\setlength{\tabcolsep}{1.5mm}{
		\begin{tabular}{@{}l|cc|cc|cc|cc|cc @{}}
			\hline
			Settings & {Content completeness}&& Deformation && Local smoothness && Aesthetic && \textbf{\textcolor{red}{Average}}\\
			 &score $\uparrow$ &&score $\uparrow$ &&score $\uparrow$ &&score $\uparrow$ &&\textbf{\textcolor{red}{score $\uparrow$}}\\
			\hline
			Scale           & \textbf{2.875}  &&0.975 &&1.878  &&1.153 && \textcolor{blue}{1.720} \\
			Crop           &   1.295 &&\textbf{2.905} &&\textbf{2.926}  && 2.355 && \textcolor{blue}{2.370} \\
			Seam-carving    &  2.829 &&0.973  &&1.000  &&1.038 && \textcolor{blue}{1.461} \\
                InGAN           &   1.662 &&0.975 &&1.007  && 0.866 && \textcolor{blue}{1.126} \\
                FR           & 1.327 &&1.812 &&1.702  &&1.535 && \textcolor{blue}{1.594} \\
                Ours           & {2.345} &&2.757  &&2.689  &&\textbf{2.538} &&\textbf{\textcolor{red}{2.582}}  \\
   
			\midrule
	\end{tabular}}
	}
	\label{tab:2}
\end{table*}

Table \ref{tab:2}  presents the performance of different methods on four subjective evaluation metrics. As shown in the table, scaling and cropping exhibit two extremes, with scaling prioritizing content completeness and cropping prioritizing shape control. In contrast, our method receives high ratings across all four evaluation metrics. Notably, when compared to Table \ref{tab:1}, scaling exhibits significant discrepancies between subjective and objective metrics in terms of key content preservation. We believe this is because the human eye has a natural interpolation ability compared to machines. Therefore, for scaling methods that uniformly delete pixels, subjective observers may not perceive strong content loss, even though objective metrics may indicate otherwise.

\begin{figure*}[!ht] 
	\centering 
	\captionsetup[subfloat]{labelsep=none,format=plain,labelformat=empty}
	\begin{minipage}[b]{0.999\linewidth} 
		\subfloat[Original image]{
			\begin{minipage}[b]{0.22\linewidth} 
				\centering
				\includegraphics[width=\linewidth]{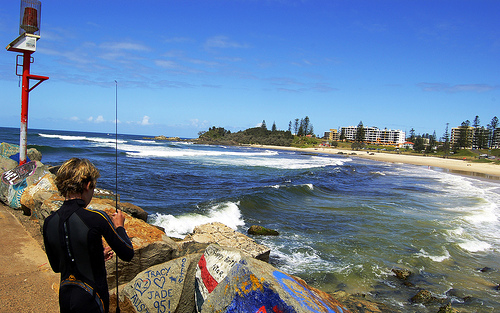}\vspace{26pt}
				\includegraphics[width=\linewidth]{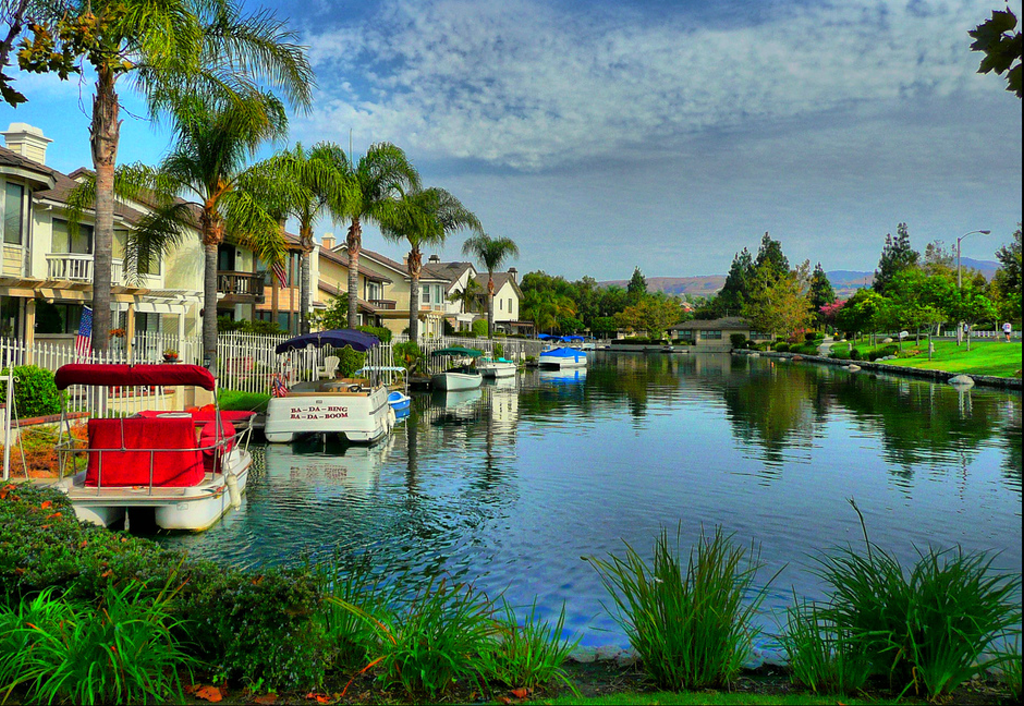}
    
			\end{minipage}
		}\hspace{-5pt}
		\subfloat[Scale]{
			\begin{minipage}[b]{0.12\linewidth}
				\centering
				\includegraphics[width=\linewidth]{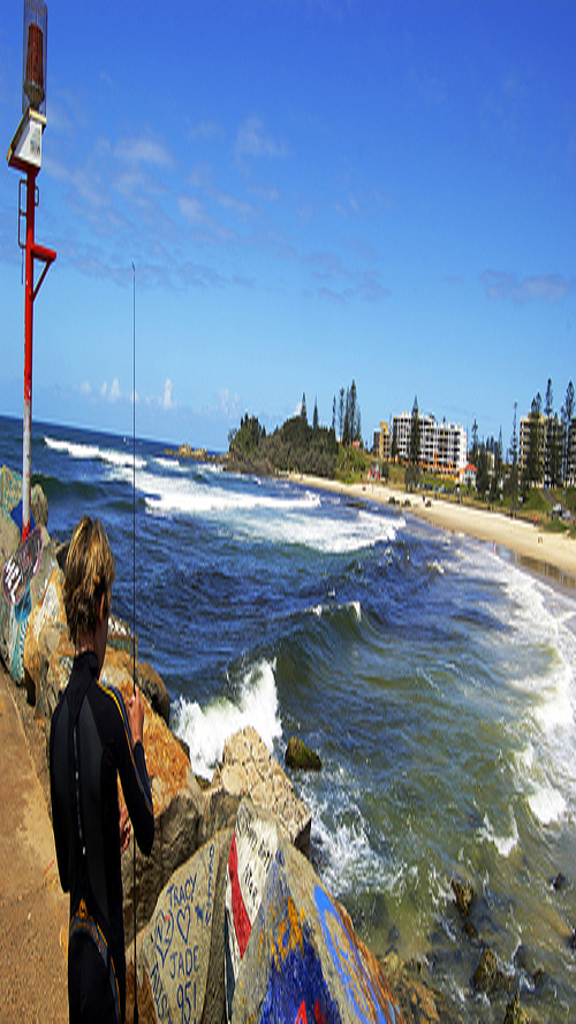}\vspace{2pt}
				\includegraphics[width=\linewidth]{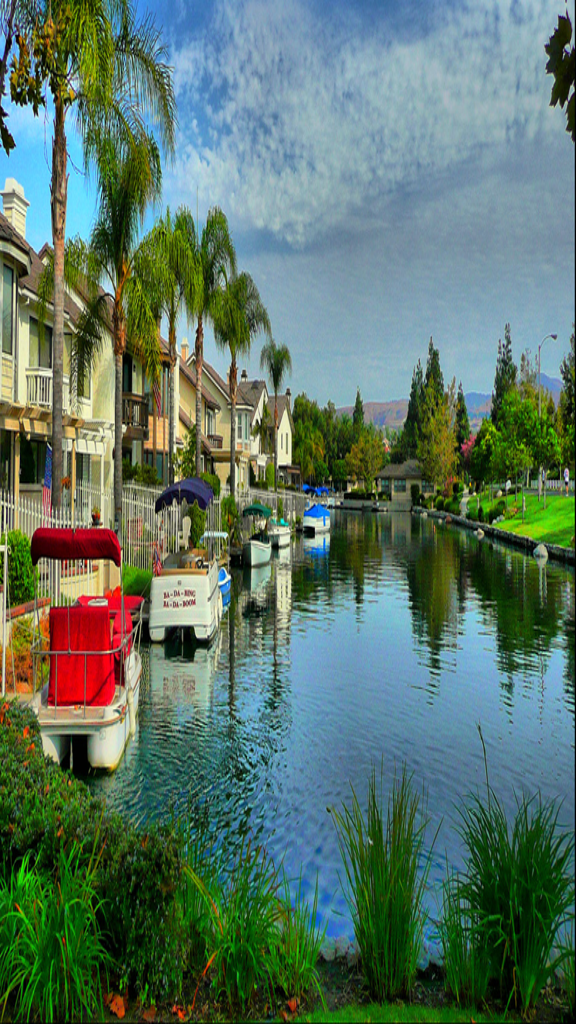}
			\end{minipage}
		}\hspace{-5pt}
		\subfloat[Crop]{
			\begin{minipage}[b]{0.12\linewidth}
				\centering
				\includegraphics[width=\linewidth]{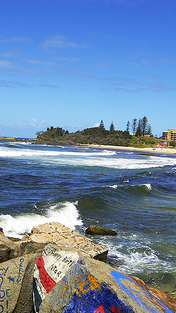}\vspace{2pt}
				\includegraphics[width=\linewidth]{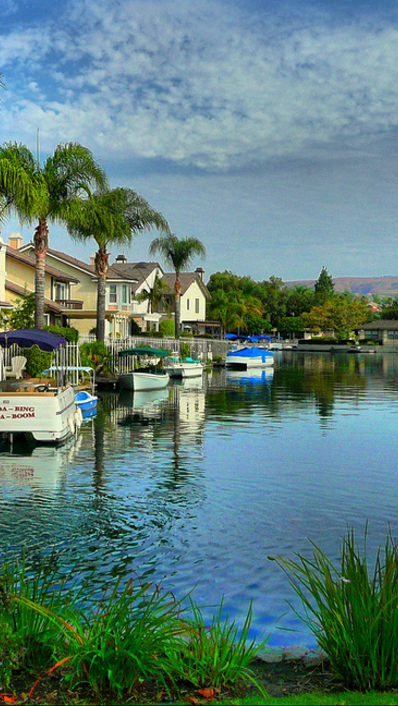}
			\end{minipage}
		}\hspace{-5pt}
		\subfloat[Seam-Carving]{
			\begin{minipage}[b]{0.12\linewidth}
				\centering
				\includegraphics[width=\linewidth]{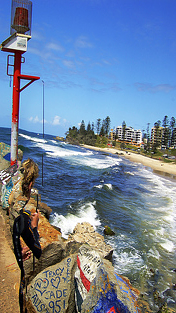}\vspace{2pt}
				\includegraphics[width=\linewidth]{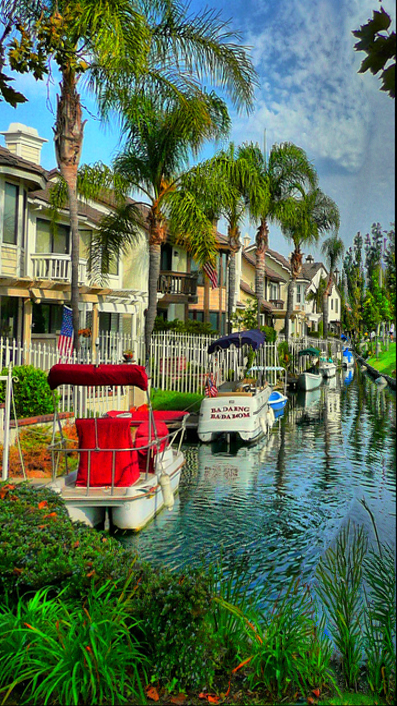}
				
			\end{minipage}
		}\hspace{-5pt}
		\subfloat[InGAN]{
			\begin{minipage}[b]{0.12\linewidth}
				\centering
				\includegraphics[width=\linewidth]{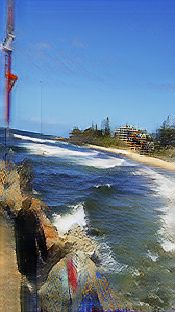}\vspace{2pt}
				\includegraphics[width=\linewidth]{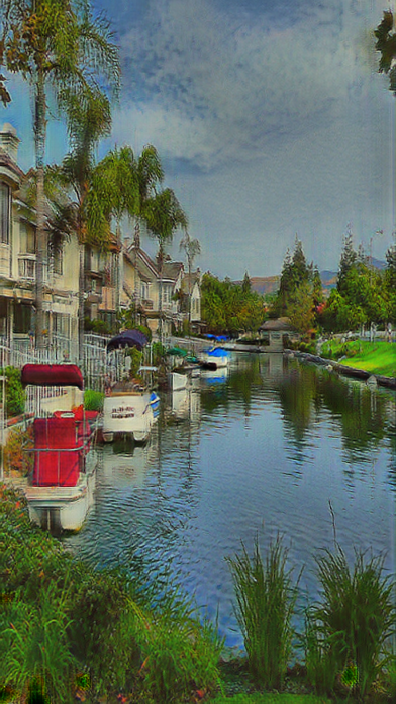}
			\end{minipage}
		}\hspace{-5pt}
		\subfloat[FR]{
			\begin{minipage}[b]{0.12\linewidth}
				\centering
				\includegraphics[width=\linewidth]{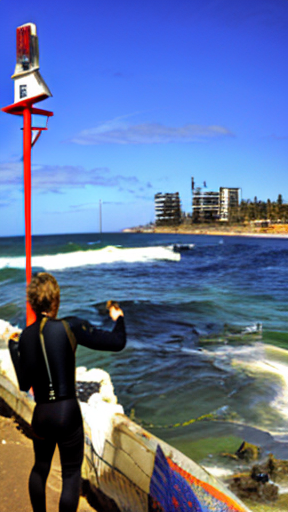}\vspace{2pt}
				\includegraphics[width=\linewidth]{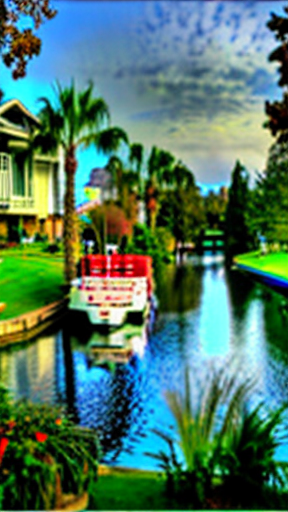}
			\end{minipage}
		}\hspace{-5pt}
		\subfloat[Ours]{
			\begin{minipage}[b]{0.12\linewidth}
				\centering
				\includegraphics[width=\linewidth]{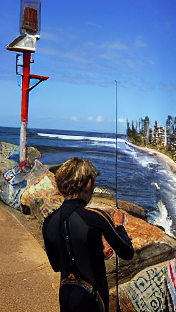}\vspace{2pt}
				\includegraphics[width=\linewidth]{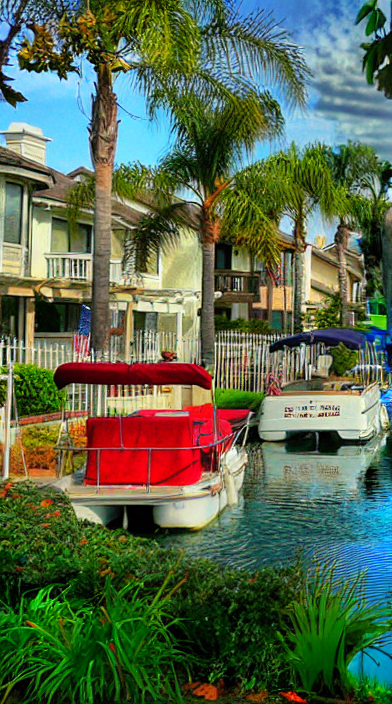}
			\end{minipage}
		}
	\end{minipage}
	\vfill
	\caption{Visual comparison to other retargeting methods on ratio 16:9.}
	\label{fig:16_9}
\end{figure*}

\begin{figure*}[!ht] 
	\centering 
	\captionsetup[subfloat]{labelsep=none,format=plain,labelformat=empty}
	\begin{minipage}[b]{0.999\linewidth} 
		\subfloat[Original image]{
			\begin{minipage}[b]{0.11\linewidth} 
				\centering
				\includegraphics[width=\linewidth]{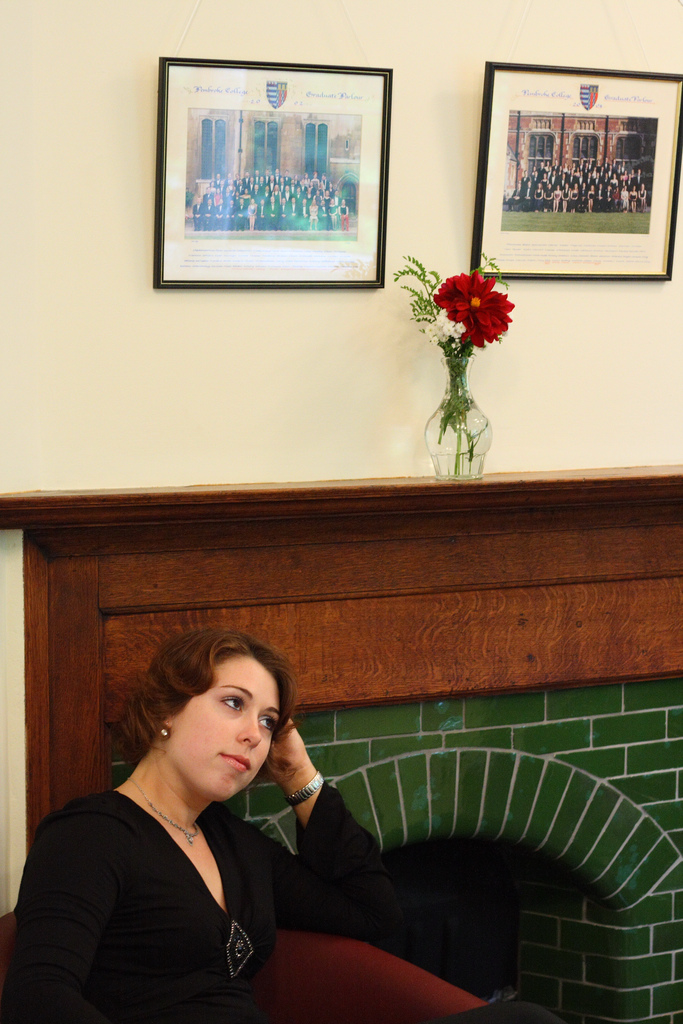}\vspace{13pt}
                    \includegraphics[width=\linewidth]{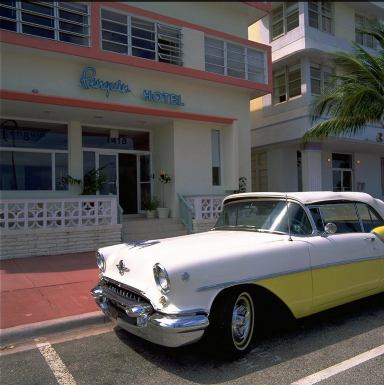}
			\end{minipage}
		}\hspace{-5pt}
		\subfloat[Scale]{
			\begin{minipage}[b]{0.14\linewidth}
				\centering
				\includegraphics[width=\linewidth]{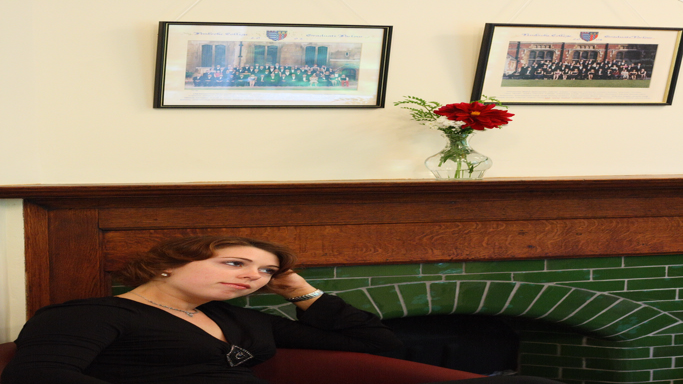}\vspace{25pt}
                    \includegraphics[width=\linewidth]{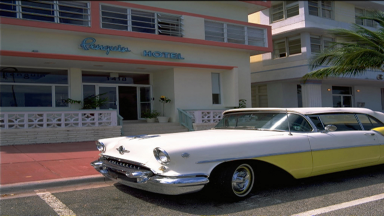}
			\end{minipage}
		}\hspace{-5pt}
		\subfloat[Crop]{
			\begin{minipage}[b]{0.14\linewidth}
				\centering
				\includegraphics[width=\linewidth]{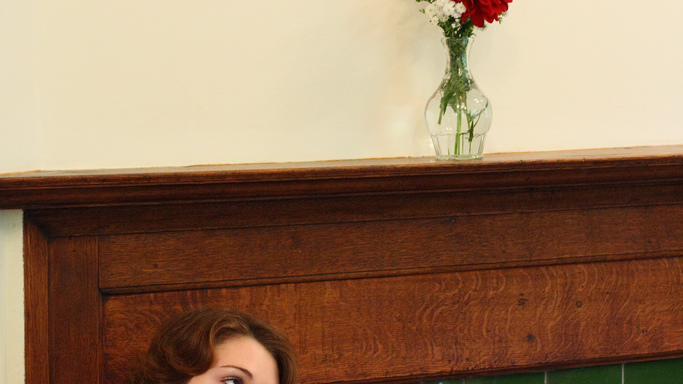}\vspace{25pt}
                    \includegraphics[width=\linewidth]{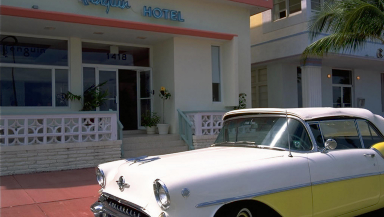}
			\end{minipage}
		}\hspace{-5pt}
		\subfloat[Seam-Carving]{
			\begin{minipage}[b]{0.14\linewidth}
				\centering
				\includegraphics[width=\linewidth]{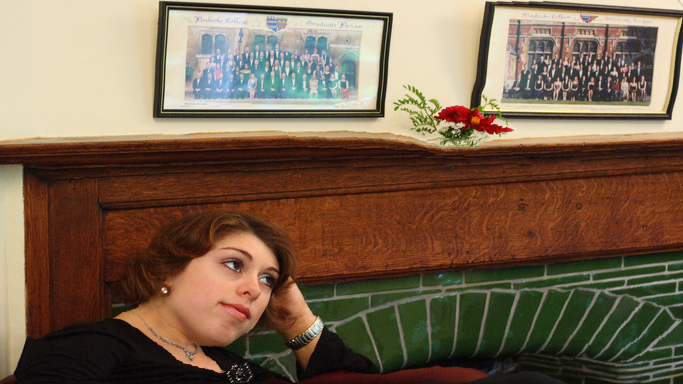}\vspace{25pt}
                    \includegraphics[width=\linewidth]{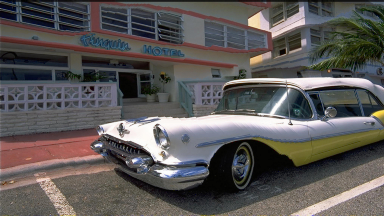}

			\end{minipage}
		}\hspace{-5pt}
            \subfloat[InGAN]{
			\begin{minipage}[b]{0.14\linewidth}
				\centering
				\includegraphics[width=\linewidth]{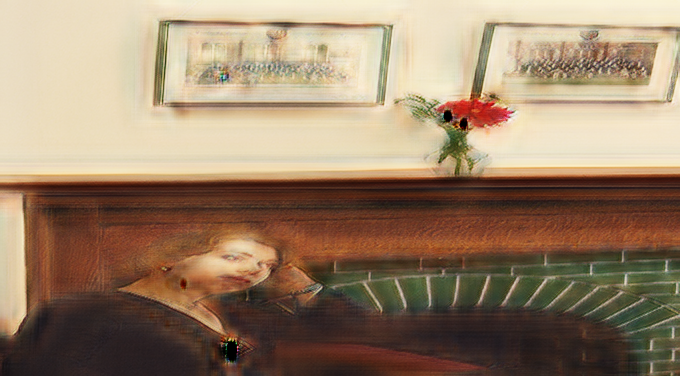}\vspace{25pt}
                    \includegraphics[width=\linewidth]{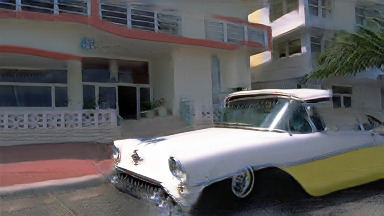}

			\end{minipage}
		}\hspace{-5pt}
            \subfloat[FR]{
			\begin{minipage}[b]{0.14\linewidth}
				\centering
				\includegraphics[width=\linewidth]{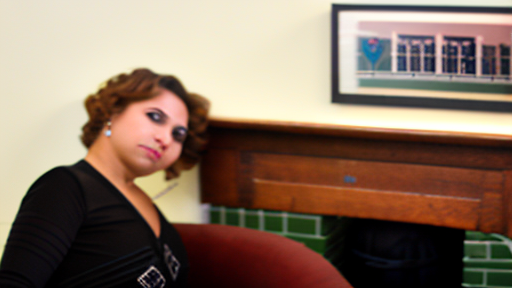}\vspace{25pt}
                    \includegraphics[width=\linewidth]{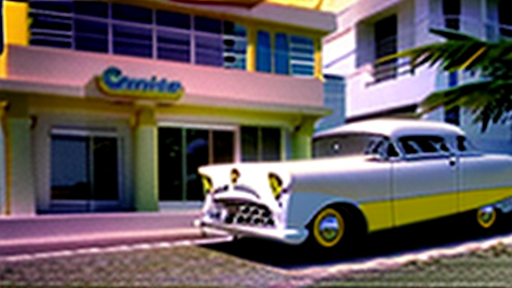}

			\end{minipage}
		}\hspace{-5pt}
		\subfloat[Ours]{
			\begin{minipage}[b]{0.14\linewidth}
				\centering
				\includegraphics[width=\linewidth]{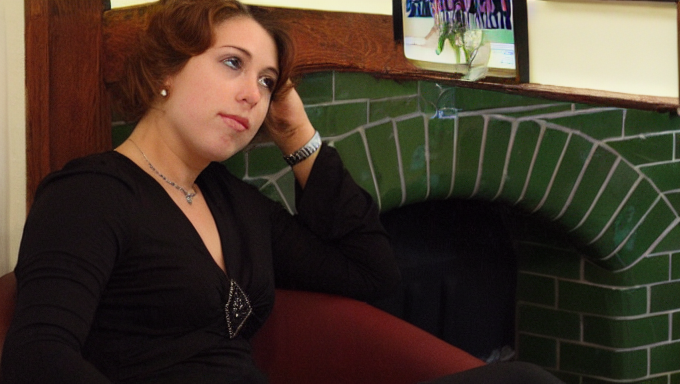}\vspace{25pt}
                    \includegraphics[width=\linewidth]{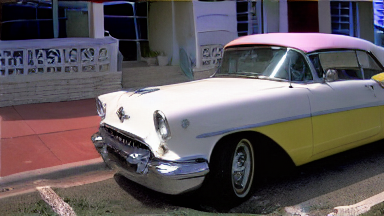}
			\end{minipage}
		}
	\end{minipage}
	\vfill
	\caption{Visual comparison to other retargeting methods on ratio 9:16.}
	\label{fig:9_16}
\end{figure*}

To qualitatively evaluate the performance of our proposed
method, we visually compare our model with other 5 retargeting methods, including scaling, cropping, seam-carving \cite{avidan2007seam}, InGAN \cite{shocher2019ingan} and full repainting (FR, which repaints the whole image with the guidance of original image using IP-Adapter \cite{ye2023ip}) on different ratios. We conduct experiments with different ratios to provide an overall assessment of each method. Figure \ref{fig:16_9} and Figure \ref{fig:9_16} illustrate the comparison of retargeting results with two extreme target ratios respectively. We can intuitively observe that most traditional methods produce inferior results due to the lack of semantic information or the oversized salient areas. They struggle to balance the trade-off between preserving key content and preventing significant object deformation. In contrast, our proposed method effectively preserves the essential content and structure of foreground objects while simultaneously maintaining harmonious and consistent background.

\subsection{Ablation Study}

In this section, we comprehensively conduct ablation experiments to verify the effectiveness of each design in our proposed model on the popular aspect ratio 16:9.

\begin{table}[ht]
    \centering
    \begin{minipage}{0.48\linewidth}
        \centering
        \caption{Ablation study of our retargeting methods on ratio 16:9.
        }
        \scalebox{1.1}{
        \setlength{\tabcolsep}{1.5mm}{
            \begin{tabular}{@{}l|c@{}}
                \hline
                Methods & SDR$ \downarrow$   \\
                \hline
                Seam-carving & 0.490  \\
                +CSC & 0.190\\
                +CSC+AR & \textbf{0.151}  \\
                \midrule
            \end{tabular}}
        }\label{tab:3}
    \end{minipage}
    \hfill
    \begin{minipage}{0.48\linewidth}
        \centering
        \caption{Comparison of background repainting (BR) and our adaptive repainting (AR) on ratio 16:9.}
        \scalebox{1.1}{
        \setlength{\tabcolsep}{1.5mm}{
            \begin{tabular}{@{}l|c@{}}
                \hline
                Methods & SDR$ \downarrow$   \\
                \hline
                Background Repainting & 0.190 \\
                Adaptive Repainting & \textbf{0.151}  \\
                \midrule
            \end{tabular}}\label{tab:4}
        }
    \end{minipage}
\end{table}

\begin{figure*}[!ht] 
	\centering 
	\captionsetup[subfloat]{labelsep=none,format=plain,labelformat=empty}
	\begin{minipage}[b]{0.92\linewidth} 
		\subfloat[Original image]{
			\begin{minipage}[b]{0.3\linewidth} 
				\centering
                \includegraphics[width=\linewidth]{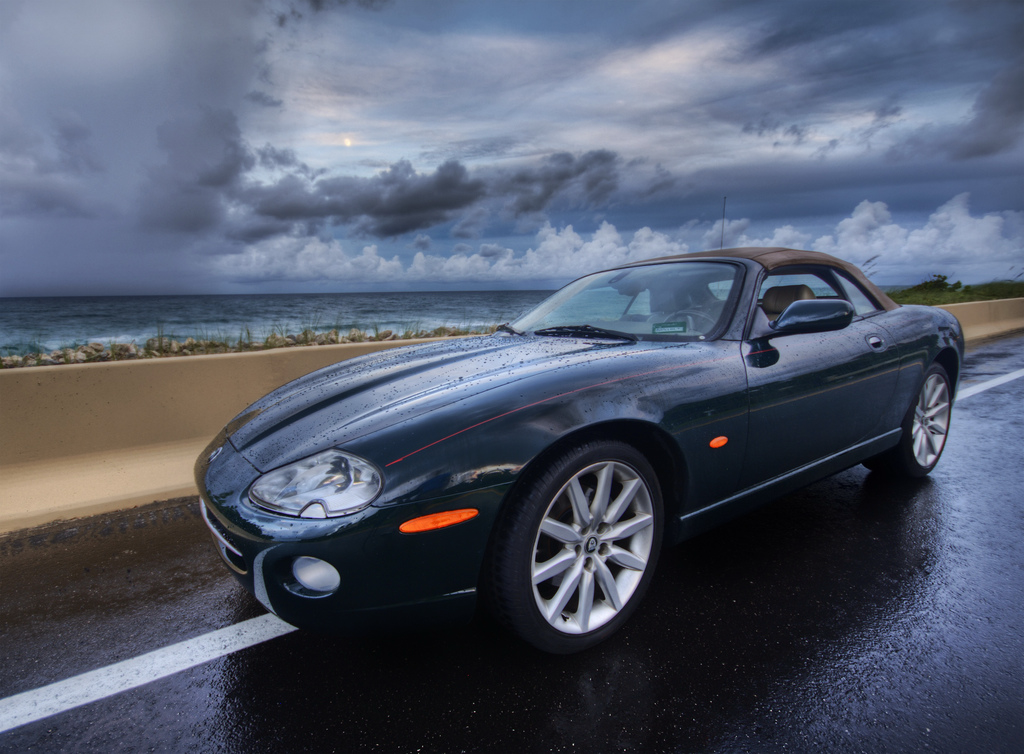}\vspace{45pt}
                \includegraphics[width=\linewidth]{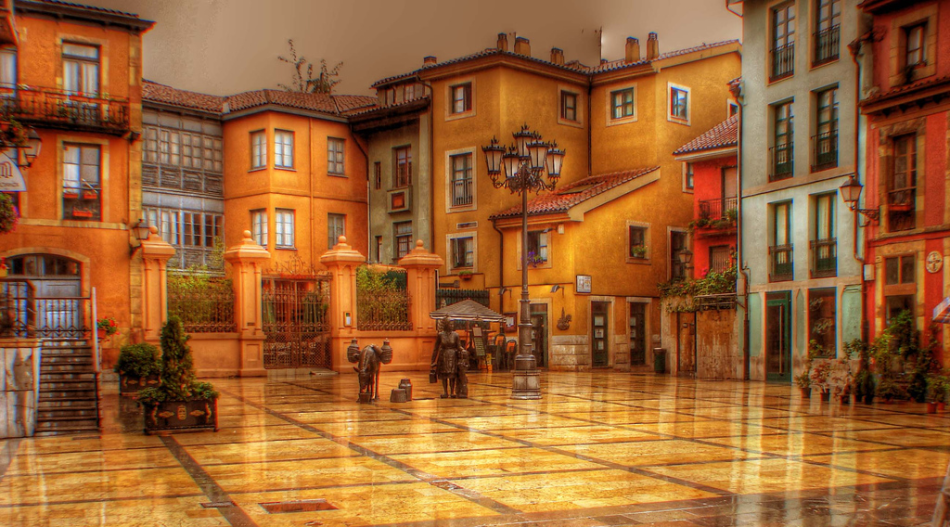}\vspace{1pt}
			\end{minipage}
		}\hspace{-5pt}
		\subfloat[Seam-carving]{
			\begin{minipage}[b]{0.16\linewidth}
				\centering
                \includegraphics[width=\linewidth]{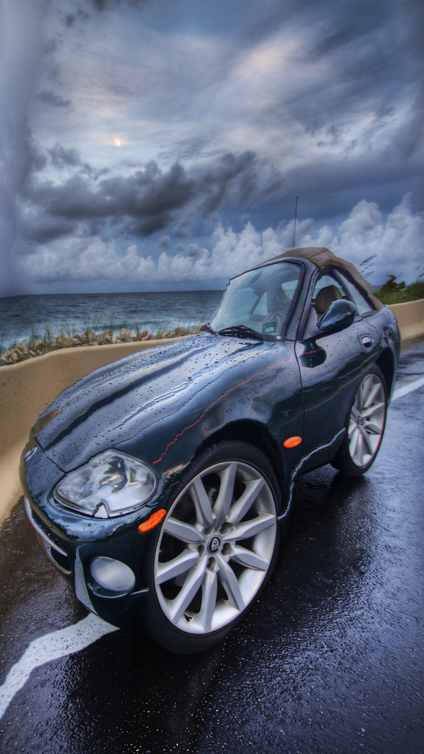}\vspace{2pt}
                \includegraphics[width=\linewidth]{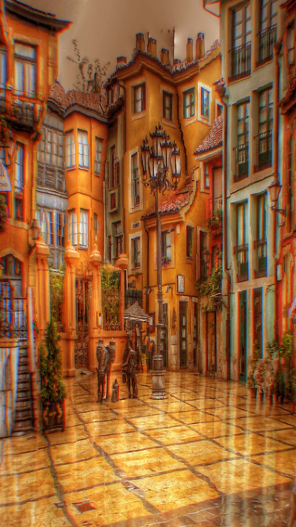}\vspace{1pt}
			\end{minipage}
		}\hspace{-5pt}
		\subfloat[+CSC]{
			\begin{minipage}[b]{0.16\linewidth}
				\centering
                \includegraphics[width=\linewidth]{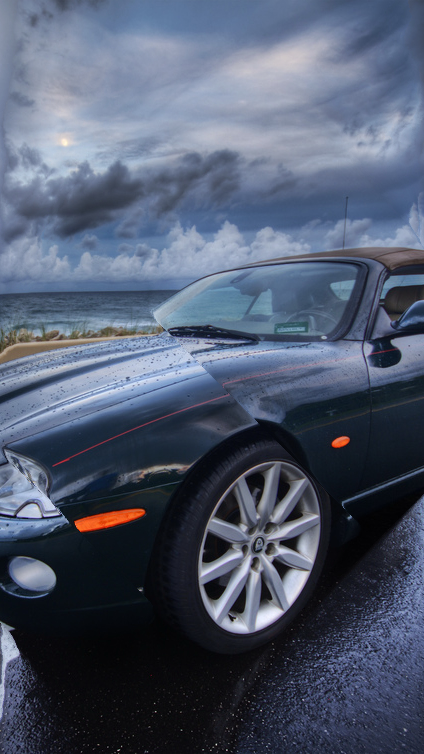}\vspace{2pt}
                \includegraphics[width=\linewidth]{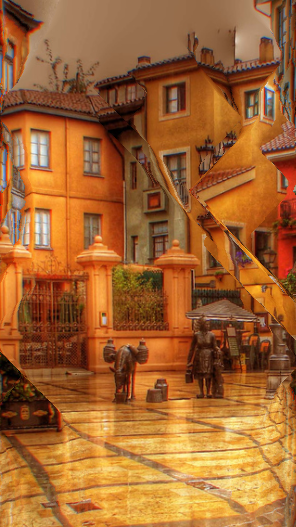}\vspace{1pt}
			\end{minipage}
		}\hspace{-5pt}
		\subfloat[ +CSC+BR]{
                \centering
			\begin{minipage}[b]{0.16\linewidth}
				\centering
                \includegraphics[width=\linewidth]{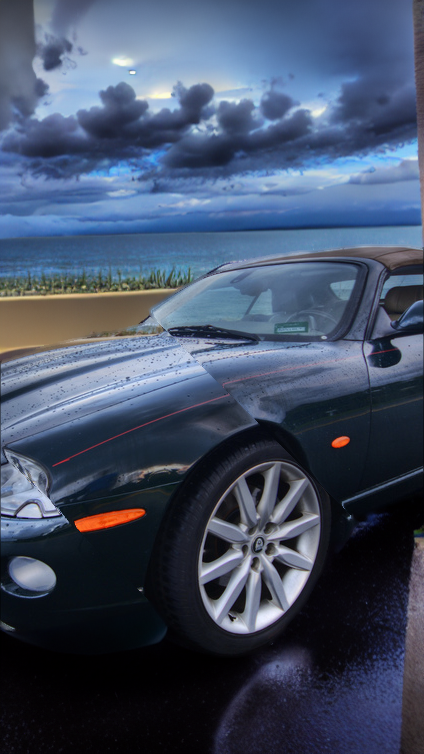}\vspace{2pt}
                \includegraphics[width=\linewidth]{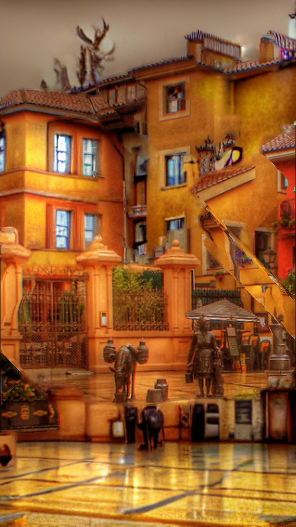}\vspace{1pt}

			\end{minipage}
		}\hspace{-5pt}
		\subfloat[+CSC+AR]{
			\begin{minipage}[b]{0.16\linewidth}
				\centering
                \includegraphics[width=\linewidth]{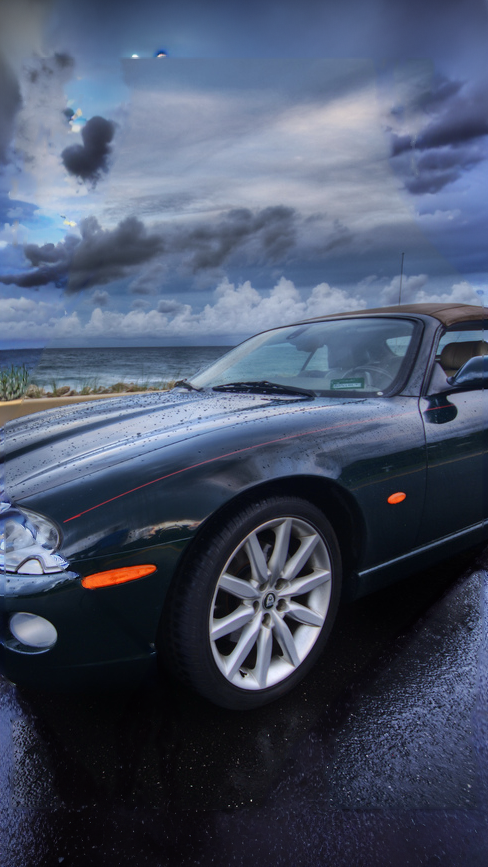}\vspace{2pt}
                \includegraphics[width=\linewidth]{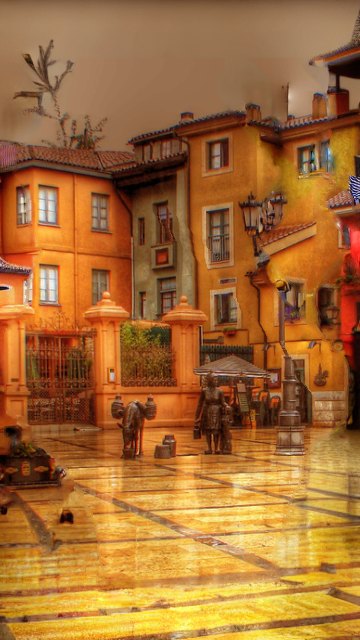}\vspace{1pt}
			\end{minipage}
		}
	\end{minipage}
	\vfill
	\caption{Visualization to demonstrate the effectiveness of each component in our method. `+CSC' denotes content-aware seam-carving in section \ref{section:csc}, `+CSC+BR' means adopt background repainting based on CSC, `+CSC+AR' means adaptive repainting in section \ref{section:repainting} based on CSC.}
	\label{fig:ablation}
\end{figure*}

\textbf{Effectiveness of content-aware seam-carving.}
As shown in Table \ref{tab:3},  content-aware seam-carving (denoted by `+CSC') significantly reduces the SDR, which means the salient objects are preserved much better. Besides, CSC can better preserve the structure of key objects, as evidenced by Figure \ref{fig:ablation}.
Different from the original seam-carving, the addition of the CSC module results in minimal deformation for the car. Also, the house with more complex patterns maintains its basic structure and avoids significant global deformation. 

\textbf{Effectiveness of adaptive repainting.}
The comparison between `+CSC' and `+CSC+AR' in Table \ref{tab:3} shows consistent improvement by the adaptive repainting module. As shown in Figure \ref{fig:ablation}, AR adaptively identifies areas with abrupt pixels for repainting and adjusts the mask according to the target aspect ratio, leading to enhanced results.

\textbf{Comparison between background repainting and adaptive repainting.} 
To further validate the advantages of our proposed adaptive repainting method, we introduce the Background Repainting (BR) strategy for comparison. BR identifies the background based on saliency maps as the region for regeneration. Table \ref{tab:4} demonstrates the advantages of our AR method in preserving salient regions, which is supported by Figure \ref{fig:ablation}. Specifically, BR is unable to address discontinuities in foreground pixels (see the car and the building in Figure \ref{fig:ablation}), and the retargeting results are constrained by the ratio (the structure of car in the first row of Figure \ref{fig:ablation} due to extreme ratio). In contrast, our AR can identify all abrupt pixel regions and adapt well to extreme ratios.

\subsection{Limitations}\label{section:limitation}

Constrained by the Stable Diffusion model, the inference speed of our method is relatively slow, averaging 7 seconds per image on the RetargetMe dataset \cite{Rubinstein2010ACS}. This may limit its real-time applicability in certain scenarios. Moreover, the repainting region generated by ARRD is not complete enough, 
as ARRD searches the local pixel displacement area without global understanding. For instance, in comparing the 'Original image' and '+CSC+AR' images in the second row of Figure \ref{fig:ablation}, several seams passing through the streetlight were removed, causing misalignment. AR only repaints the pixels around the deleted seams instead of the entire streetlight, resulting in a streetlight that remains misaligned in the generated image.

\section{Conclusion}\label{section:conclusion}

Our paper introduces a new image retargeting model called PruneRepaint. This model is content-aware and adaptive, allowing it to work with any target ratio. The content-aware seam-carving method protects important semantic regions, while the adaptive repainting module helps to maintain visual quality even after pixels are deleted. Through extensive experiments, we have demonstrated the effectiveness of our design and the advantages of using PruneRepaint for image retargeting.

\section*{Acknowledgments}

The work was supported in part by the National
Natural Science Foundation of China (NSFC) under Grant 62102083; in
part by the Natural Science Foundation of Jiangsu Province under Grant
BK20210222; in part by the National Natural Science Foundation of China
(NSFC) under Grant 62261160576, Grant 62203024, and Grant 92167102;
and in part by the Research and Development Program of Beijing Municipal
Education Commission under Grant KM202310005027.

\bibliographystyle{plainnat}
\bibliography{neurips_2024.bib}

\begin{thebibliography}{39}
\providecommand{\natexlab}[1]{#1}
\providecommand{\url}[1]{\texttt{#1}}
\expandafter\ifx\csname urlstyle\endcsname\relax
  \providecommand{\doi}[1]{doi: #1}\else
  \providecommand{\doi}{doi: \begingroup \urlstyle{rm}\Url}\fi

\bibitem[Avidan and Shamir(2007)]{avidan2007seam}
Shai Avidan and Ariel Shamir.
\newblock Seam carving for content-aware image resizing.
\newblock In \emph{ACMSIGGRAPH}, page 10–es, New York, NY, USA, 2007.
  Association for Computing Machinery.
\newblock ISBN 9781450378369.
\newblock \doi{10.1145/1275808.1276390}.
\newblock URL \url{https://doi.org/10.1145/1275808.1276390}.

\bibitem[Canny(1986)]{canny1986computational}
John Canny.
\newblock A computational approach to edge detection.
\newblock \emph{TPAMI}, \penalty0 (6):\penalty0 679--698, 1986.

\bibitem[Cho et~al.(2017)Cho, Park, Oh, Tai, and So~Kweon]{cho2017weakly}
Donghyeon Cho, Jinsun Park, Tae-Hyun Oh, Yu-Wing Tai, and In~So~Kweon.
\newblock Weakly-and self-supervised learning for content-aware deep image
  retargeting.
\newblock In \emph{ICCV}, pages 4558--4567, 2017.

\bibitem[Dickman et~al.(2023)Dickman, Diefenbach, Burlick, and
  Stockton]{dickman2023smart}
Elliot Dickman, Paul Diefenbach, Matthew Burlick, and Mark Stockton.
\newblock Smart scaling: A hybrid deep-learning approach to content-aware image
  retargeting.
\newblock In \emph{ACMSIGGRAPH}, pages 1--2. 2023.

\bibitem[Dong et~al.(2009)Dong, Zhou, Paul, and Zhang]{dong2009optimized}
Weiming Dong, Ning Zhou, Jean-Claude Paul, and Xiaopeng Zhang.
\newblock Optimized image resizing using seam carving and scaling.
\newblock \emph{TOG}, 28\penalty0 (5):\penalty0 1--10, 2009.

\bibitem[Duda et~al.(1973)Duda, Hart, and Stork]{duda1973pattern}
Richard~O Duda, Peter~E Hart, and David~G Stork.
\newblock \emph{Pattern classification and scene analysis}, volume~3.
\newblock Wiley New York, 1973.

\bibitem[Dy et~al.(2023)Dy, Virtusio, Tan, Lin, Ilao, Chen, and
  Hua]{dy2023mcgan}
Jilyan~Bianca Dy, John~Jethro Virtusio, Daniel~Stanley Tan, Yong-Xiang Lin,
  Joel Ilao, Yung-Yao Chen, and Kai-Lung Hua.
\newblock Mcgan: mask controlled generative adversarial network for image
  retargeting.
\newblock \emph{Neural. Comput. Appl}, 35\penalty0 (14):\penalty0 10497--10509,
  2023.

\bibitem[Fan et~al.(2021)Fan, Lei, Liang, Fang, Cao, and
  Ling]{fan2021unsupervised}
Xiaoting Fan, Jianjun Lei, Jie Liang, Yuming Fang, Xiaochun Cao, and Nam Ling.
\newblock Unsupervised stereoscopic image retargeting via view synthesis and
  stereo cycle consistency losses.
\newblock \emph{Neurocomputing}, 447:\penalty0 161--171, 2021.

\bibitem[Fan et~al.(2024)Fan, Zhang, Sun, Xiao, and
  Durrani]{fan2024comprehensive}
Xiaoting Fan, Zhong Zhang, Long Sun, Baihua Xiao, and Tariq~S Durrani.
\newblock A comprehensive review of image retargeting.
\newblock \emph{Neurocomputing}, page 127416, 2024.

\bibitem[Goodfellow et~al.(2020)Goodfellow, Pouget-Abadie, Mirza, Xu,
  Warde-Farley, Ozair, Courville, and Bengio]{goodfellow2020generative}
Ian Goodfellow, Jean Pouget-Abadie, Mehdi Mirza, Bing Xu, David Warde-Farley,
  Sherjil Ozair, Aaron Courville, and Yoshua Bengio.
\newblock Generative adversarial networks.
\newblock \emph{COMMUN ACM}, 63\penalty0 (11):\penalty0 139--144, 2020.

\bibitem[Ho et~al.(2020)Ho, Jain, and Abbeel]{ho2020denoising}
Jonathan Ho, Ajay Jain, and Pieter Abbeel.
\newblock Denoising diffusion probabilistic models.
\newblock \emph{NeuriPS}, 33:\penalty0 6840--6851, 2020.

\bibitem[Houlsby et~al.(2019)Houlsby, Giurgiu, Jastrzebski, Morrone,
  De~Laroussilhe, Gesmundo, Attariyan, and Gelly]{houlsby2019parameter}
Neil Houlsby, Andrei Giurgiu, Stanislaw Jastrzebski, Bruna Morrone, Quentin
  De~Laroussilhe, Andrea Gesmundo, Mona Attariyan, and Sylvain Gelly.
\newblock Parameter-efficient transfer learning for nlp.
\newblock In \emph{ICML}, pages 2790--2799. PMLR, 2019.

\bibitem[Hsu et~al.(2014)Hsu, Lin, Fang, and Lin]{hsu2014objective}
Chih-Chung Hsu, Chia-Wen Lin, Yuming Fang, and Weisi Lin.
\newblock Objective quality assessment for image retargeting based on
  perceptual geometric distortion and information loss.
\newblock \emph{IEEE J. Sel. Top. Signal Process}, 8\penalty0 (3):\penalty0
  377--389, 2014.

\bibitem[Kajiura et~al.(2020)Kajiura, Kosugi, Wang, and
  Yamasaki]{kajiura2020self}
Nobukatsu Kajiura, Satoshi Kosugi, Xueting Wang, and Toshihiko Yamasaki.
\newblock Self-play reinforcement learning for fast image retargeting.
\newblock In \emph{ACM Multimedia}, 2020.

\bibitem[LeCun et~al.(1998)LeCun, Bottou, Bengio, and
  Haffner]{lecun1998gradient}
Yann LeCun, L{\'e}on Bottou, Yoshua Bengio, and Patrick Haffner.
\newblock Gradient-based learning applied to document recognition.
\newblock \emph{Proc. IEEE}, 86\penalty0 (11):\penalty0 2278--2324, 1998.

\bibitem[Lin et~al.(2019)Lin, Zhou, and Chen]{lin2019deepir}
Jianxin Lin, Tiankuang Zhou, and Zhibo Chen.
\newblock Deepir: A deep semantics driven framework for image retargeting.
\newblock In \emph{ICMEW}, pages 54--59. IEEE, 2019.

\bibitem[Liu et~al.(2021)Liu, Zhang, Wan, Shao, and Han]{liu2021visual}
Nian Liu, Ni~Zhang, Kaiyuan Wan, Ling Shao, and Junwei Han.
\newblock Visual saliency transformer.
\newblock In \emph{ICCV}, pages 4722--4732, 2021.

\bibitem[Liu et~al.(2011)Liu, Luo, Xuan, Chen, and Fu]{liu2011image}
Yong-Jin Liu, Xi~Luo, Yu-Ming Xuan, Wen-Feng Chen, and Xiao-Lan Fu.
\newblock Image retargeting quality assessment.
\newblock In \emph{Computer Graphics Forum}, volume~30, pages 583--592. Wiley
  Online Library, 2011.

\bibitem[Lugmayr et~al.(2022)Lugmayr, Danelljan, Romero, Yu, Timofte, and
  Van~Gool]{lugmayr2022repaint}
Andreas Lugmayr, Martin Danelljan, Andres Romero, Fisher Yu, Radu Timofte, and
  Luc Van~Gool.
\newblock Repaint: Inpainting using denoising diffusion probabilistic models.
\newblock In \emph{CVPR}, pages 11461--11471, 2022.

\bibitem[Ma et~al.(2012)Ma, Lin, Deng, and Ngan]{ma2012image}
Lin Ma, Weisi Lin, Chenwei Deng, and King~Ngi Ngan.
\newblock Image retargeting quality assessment: A study of subjective scores
  and objective metrics.
\newblock \emph{IEEE J. Sel. Top. Signal Process}, 6\penalty0 (6):\penalty0
  626--639, 2012.

\bibitem[Manjunath et~al.(2001)Manjunath, Ohm, Vasudevan, and
  Yamada]{manjunath2001color}
Bangalore~S Manjunath, J-R Ohm, Vinod~V Vasudevan, and Akio Yamada.
\newblock Color and texture descriptors.
\newblock \emph{TCSVT}, 11\penalty0 (6):\penalty0 703--715, 2001.

\bibitem[Mei et~al.(2021)Mei, Guo, Sun, Pan, and Zhang]{mei2021deep}
Yijing Mei, Xiaojie Guo, Di~Sun, Gang Pan, and Jiawan Zhang.
\newblock Deep supervised image retargeting.
\newblock In \emph{ICME}, pages 1--6. IEEE, 2021.

\bibitem[Metz et~al.(2016)Metz, Poole, Pfau, and
  Sohl-Dickstein]{metz2016unrolled}
Luke Metz, Ben Poole, David Pfau, and Jascha Sohl-Dickstein.
\newblock Unrolled generative adversarial networks.
\newblock In \emph{ICLR}, 2016.

\bibitem[Podell et~al.(2023)Podell, English, Lacey, Blattmann, Dockhorn,
  M{\"u}ller, Penna, and Rombach]{podell2023sdxl}
Dustin Podell, Zion English, Kyle Lacey, Andreas Blattmann, Tim Dockhorn, Jonas
  M{\"u}ller, Joe Penna, and Robin Rombach.
\newblock Sdxl: Improving latent diffusion models for high-resolution image
  synthesis.
\newblock \emph{arXiv preprint arXiv:2307.01952}, 2023.

\bibitem[Radford et~al.(2021)Radford, Kim, Hallacy, Ramesh, Goh, Agarwal,
  Sastry, Askell, Mishkin, Clark, et~al.]{radford2021learning}
Alec Radford, Jong~Wook Kim, Chris Hallacy, Aditya Ramesh, Gabriel Goh,
  Sandhini Agarwal, Girish Sastry, Amanda Askell, Pamela Mishkin, Jack Clark,
  et~al.
\newblock Learning transferable visual models from natural language
  supervision.
\newblock In \emph{ICML}, pages 8748--8763. PMLR, 2021.

\bibitem[Rombach et~al.(2022)Rombach, Blattmann, Lorenz, Esser, and
  Ommer]{rombach2022high}
Robin Rombach, Andreas Blattmann, Dominik Lorenz, Patrick Esser, and Bj{\"o}rn
  Ommer.
\newblock High-resolution image synthesis with latent diffusion models.
\newblock In \emph{CVPR}, pages 10684--10695, 2022.

\bibitem[Rubinstein et~al.(2008)Rubinstein, Shamir, and
  Avidan]{rubinstein2008improved}
Michael Rubinstein, Ariel Shamir, and Shai Avidan.
\newblock Improved seam carving for video retargeting.
\newblock \emph{TOG}, 27\penalty0 (3):\penalty0 1--9, 2008.

\bibitem[Rubinstein et~al.(2010)Rubinstein, Gutierrez, Sorkine-Hornung, and
  Shamir]{Rubinstein2010ACS}
Michael Rubinstein, Diego Gutierrez, Olga Sorkine-Hornung, and Ariel Shamir.
\newblock A comparative study of image retargeting.
\newblock \emph{ACMSIGGRAPH}, 2010.
\newblock URL \url{https://api.semanticscholar.org/CorpusID:3332468}.

\bibitem[Santella et~al.(2006)Santella, Agrawala, DeCarlo, Salesin, and
  Cohen]{santella2006gaze}
Anthony Santella, Maneesh Agrawala, Doug DeCarlo, David Salesin, and Michael
  Cohen.
\newblock Gaze-based interaction for semi-automatic photo cropping.
\newblock In \emph{SIGCHI}, pages 771--780, 2006.

\bibitem[Shaham et~al.(2019)Shaham, Dekel, and Michaeli]{shaham2019singan}
Tamar~Rott Shaham, Tali Dekel, and Tomer Michaeli.
\newblock Singan: Learning a generative model from a single natural image.
\newblock In \emph{ICCV}, pages 4570--4580, 2019.

\bibitem[Shocher et~al.(2019)Shocher, Bagon, Isola, and
  Irani]{shocher2019ingan}
Assaf Shocher, Shai Bagon, Phillip Isola, and Michal Irani.
\newblock Ingan: Capturing and retargeting the" dna" of a natural image.
\newblock In \emph{ICCV}, pages 4492--4501, 2019.

\bibitem[Simonyan and Zisserman(2014)]{simonyan2014very}
Karen Simonyan and Andrew Zisserman.
\newblock Very deep convolutional networks for large-scale image recognition.
\newblock \emph{arXiv preprint arXiv:1409.1556}, 2014.

\bibitem[Song et~al.(2020)Song, Meng, and Ermon]{song2020denoising}
Jiaming Song, Chenlin Meng, and Stefano Ermon.
\newblock Denoising diffusion implicit models.
\newblock \emph{arXiv preprint arXiv:2010.02502}, 2020.

\bibitem[Valdez-Balderas et~al.(2021)Valdez-Balderas, Muraveynyk, and
  Smith]{valdez2021fast}
Daniel Valdez-Balderas, Oleg Muraveynyk, and Timothy Smith.
\newblock Fast hybrid image retargeting.
\newblock In \emph{ICIP}, pages 1849--1853. IEEE, 2021.

\bibitem[Vaquero et~al.(2010)Vaquero, Turk, Pulli, Tico, and
  Gelfand]{vaquero2010survey}
Daniel Vaquero, Matthew Turk, Kari Pulli, Marius Tico, and Natasha Gelfand.
\newblock A survey of image retargeting techniques.
\newblock In \emph{ADIP XXXIII}, volume 7798, pages 328--342. SPIE, 2010.

\bibitem[Ye et~al.(2023)Ye, Zhang, Liu, Han, and Yang]{ye2023ip}
Hu~Ye, Jun Zhang, Sibo Liu, Xiao Han, and Wei Yang.
\newblock Ip-adapter: Text compatible image prompt adapter for text-to-image
  diffusion models.
\newblock \emph{arXiv preprint arXiv:2308.06721}, 2023.

\bibitem[Zhang et~al.(2023)Zhang, Rao, and Agrawala]{zhang2023adding}
Lvmin Zhang, Anyi Rao, and Maneesh Agrawala.
\newblock Adding conditional control to text-to-image diffusion models.
\newblock In \emph{ICCV}, pages 3836--3847, 2023.

\bibitem[Zhang et~al.(2005)Zhang, Zhang, Sun, Feng, and Ma]{zhang2005auto}
Mingju Zhang, Lei Zhang, Yanfeng Sun, Lin Feng, and Weiying Ma.
\newblock Auto cropping for digital photographs.
\newblock In \emph{ICME}, pages 4--pp. IEEE, 2005.

\bibitem[Zhou et~al.(2020)Zhou, Chen, and Li]{zhou2020weakly}
Ya~Zhou, Zhibo Chen, and Weiping Li.
\newblock Weakly supervised reinforced multi-operator image retargeting.
\newblock \emph{TCSVT}, 31\penalty0 (1):\penalty0 126--139, 2020.

\end{thebibliography}

\end{document}